\documentclass{article}
\usepackage[preprint]{corl_2026}
\usepackage{booktabs}
\usepackage{array}
\usepackage{microtype}

\title{When Human Corrections Travel: Safety in Federated VLA Adaptation}
\author{
  \small
  Vigneshwaran Subbaraju\\
  A*STAR, Singapore\\
  \texttt{Vigneshwaran\_Subbaraju}\\
  \texttt{@a-star.edu.sg}
  \And
  Obaidullah Zaland\\
  Ume\r{a} University, Sweden\\
  \texttt{ozaland@cs.umu.se}
  \And
  Monowar Bhuyan\\
  Ume\r{a} University, Sweden\\
  \texttt{monowar@cs.umu.se}
}

\begin{document}
\maketitle

\begin{abstract}
Vision-language-action (VLA) policies can adapt to local human intervention, while federated learning can combine distributed robot experience without pooling raw records. We examine their prospective intersection: sharing adaptations derived from diverse human corrections across tasks, environments, people, and robot embodiments. Corrections differ in modality, safety semantics(urgency, behavioral scope, etc.), and privacy sensitivity. A shared update that improves mean task success may suppress a rare safety intervention or produce an unsafe response on another robot. We argue for three requirements: preserve the context and uncertainty of multimodal corrections when deciding what to share; assess candidate updates in each receiving robot's closed loop before deployment; and evaluate leakage alongside retention of safety-critical evidence. Immediate, legible defensive behavior remains necessary during human interaction. We offer counterexamples and a falsifiable evaluation agenda, not experimental results.
\end{abstract}


\section{Introduction}

Vision-language-action (VLA) models connect visual observations and language instructions to robot actions~\citep{zitkovich2023rt2}. Local robot policies can learn from human intervention and online correction~\citep{jiang2025transic,chen2025mereq}; UniSteer specifically adapts a VLA using human takeover actions~\citep{lu2026unisteer}. In parallel, FedVLA and ForgeVLA show ways to learn from distributed VLA task data without centralizing raw interaction records~\citep{miao2025fedvla,zhou2026forgevla}. A recent study of federated VLA fine-tuning reports that cross-site heterogeneity is more pronounced on physical robots than in its simulations~\citep{duan2026empirical}. These strands motivate a prospective setting: robots learn locally from human corrections, then selectively share resulting adaptations. However, these federated systems do not establish the safety of federating multimodal corrections.

Such feedback is heterogeneous in ways that task labels do not capture \ref{tab:heterogeneity}. People may signal a similar intended change by speech, gesture, guided motion, or teleoperation, with different urgency and precision. Consider two sites where a human helps a robot insert a panel. At one site, an operator standing beside the fixture points to a narrow gap and says \emph{bring it through here}; the pointing gesture performed near the intended path clearly resolves an ambiguous verbal reference. At another site, an operator farther from the fixture gives a detailed verbal description or draws the desired trajectory on a camera image. The reliability of pointing changes with distance, while a drawn 2D route alone may leave depth, contact force, and handover timing unspecified \citep{weerakoon2020gesture,mehta2025l2d2}. Each correction may improve its local policy, but their learned effects are not interchangeable. At a receiving site with a rotated fixture, a different gripper, or a different human hand position, a shared update could reduce clearance or release a jointly held panel too early. A learned motion may also behave differently with another gripper, compliance controller, or action decoder. Federated optimization can encounter client distribution shift and conflicting updates; in embodied systems, a regression can increase contact force, delay a yield, or leave a shared load unsupported. A successful average trajectory offers little evidence about a rare, hazardous interaction. Updates can also reveal private training information even when raw observations remain local~\citep{zhao2024leakage}.

\textbf{Our question.} \emph{What and how may robots learn from diverse, multimodal human corrections across sites, and when may that learning safely change another robot's behavior?} Consider a person and robot jointly inserting a panel. A spoken \emph{wait}, a raised hand, and physical guidance might all signal a needed hold, but they may refer to different hazards or changes in task goal. The source robot can learn useful timing; but the same correction on a receiving robot with a different grip and fixture could make it release too early. We argue that corrective experience should travel only with a justified scope, receiving-site safety assessment, and an explicit privacy risk assessment. We develop three positions and an evaluation agenda rather than a new algorithm or empirical result.

\section{Diverse corrections require scoped transfer}

Each site adapts a pretrained VLA from locally collected interactions, which may include speech, pointing, drawn routes, physical guidance, and overrides. Sites may share selected learned updates while retaining robot-specific perception and control components. Before an update is transferred, its source context and intended behavioral change need to be identified: did the person specify a goal, a route, a constraint, or an immediate intervention? The receiving robot must then assess whether the learned change has the intended physical effect in its own task and embodiment.
A correction can target the current action, future trajectory, goal, safety constraint, or operator preference. Its channel alone does not determine its meaning. Even the same command, such as \emph{move left}, can mean avoiding a hand, clearing a fixture, or satisfying a local preference.

\begin{table}[h]
\centering
\small
\caption{Sources of heterogeneity and possible safety failures in corrective transfer.}
\label{tab:heterogeneity}
\setlength{\tabcolsep}{4pt}
\begin{tabular}{@{}p{0.19\linewidth}p{0.36\linewidth}p{0.34\linewidth}@{}}
\toprule
Variation & What may be lost in a shared update & Possible receiving-site failure \\
\midrule
Feedback channel & Urgency or precision of a spoken veto, gesture, or guided motion. & Delayed or incorrect intervention response. \\
Task and scene & Hazard location, object support, or agreed interaction protocol. & Local preference treated as a general safety rule. \\
Embodiment & Gripper, reachable states, compliance, and action representation. & Unintended contact or release of a shared load. \\
Client frequency & A rare override among many routine successes. & Safety-critical behavior attenuated by aggregation. \\
\bottomrule
\end{tabular}
\end{table}

\textbf{Position 1: share corrections according to behavioral scope, not just parameter compatibility.} A proposed update should carry or be evaluated against the source modality, relevant task and human context, intended change, uncertainty, and permitted sharing scope. This is a requirement for accountable adaptation. A guided trajectory may be informative about one robot's geometry while a spoken \emph{stop} may impose an immediate constraint with little reusable trajectory data. Converting takeover actions into training targets, as UniSteer does~\citep{lu2026unisteer}, does not by itself resolve whether an intervention was a safety veto, task correction, or preference. Multimodal federated learning addresses clients that hold different or missing modalities, mainly by aligning or ensembling representations~\citep{feng2023fedmultimodal,yu2023multimodal,ouyang2023harmony}. Such alignment can make a spoken \emph{stop} and a raised hand close in embedding space without establishing that they impose the same constraint. When they don't, retaining the correction locally is a valid decision.


\section{Transfer safety is a receiving-site property}

A feasible design shares some policy components while retaining local action mappings; FedVLA uses personalized components~\citep{miao2025fedvla}, and Align-Then-stEer aligns distinct action spaces for cross-embodiment adaptation~\citep{zhang2026ate}. Neither parameter separation nor action alignment proves that a correction will transfer safely. The source and receiver may differ in task, hazard geometry, human proximity, dynamics, and intervention protocol. The unit to assess is the receiver's adapted closed-loop system: perception, policy, action mapping, physical safeguards, and human responses.

\textbf{Position 2: a shared update needs a recipient-specific safety gate.} Before deployment, a receiver could stage the candidate update and compare it with its current policy from matched local states. An action-conditioned world model could simulate both policies under candidate hazards. SafeDojo uses imagined VLA trajectories to estimate reward and collision cost for safe policy optimization~\citep{tang2026safedojo}; using such rollouts for federated update acceptance is our proposal, not a result of that paper. The comparison should include contact and force, load support, task progress, interruption response, and legible yielding. Simulation alone misses unmodeled dynamics and human reactions, so false acceptance must be measured against local constraint tests and, where appropriate, supervised physical trials. A receiving robot should reject or roll back an update that worsens its safety cases. Local rejection does not remove a harmful contribution already in the shared model; that requires federated unlearning~\citep{liu2021federaser,halimi2022federated}. Because approximate unlearning can leave residual behavior, removal should be verified by the same receiver-side safety tests rather than by parameter distance alone. Report the worst-affected receiver as well as mean task success; independent runtime safeguards remain essential~\citep{tolle2025safe}.

\textbf{A transfer counterexample.} Robot A receives physical guidance away from a person's hand during panel insertion. Robot B completes many uncomplicated insertions and receives only success reports. Robot C, with another gripper and work height, receives a verbal request to slow down, which might be a local preference or a needed contact constraint. An aggregate could improve mean insertion success while weakening A's avoidance response or exporting C's timing to a robot for which it is unsafe. 
The task label and parameter shape do not establish that the underlying corrective experiences are interchangeable.

Faulty interpretations and malicious or corrupted client updates create further integrity risks. Privacy mechanisms that hide individual updates may make anomalous contributors harder to inspect; model poisoning work demonstrates this tension in federated learning~\citep{bagdasaryan2020backdoor}. Receiving-site validation should therefore test the behavioral consequence, regardless of whether the aggregate appears statistically typical.

\section{Privacy must preserve safety-critical evidence}

Corrections may reveal a goal, a workspace detail, habitual action, or a person's difficulty. Storing raw feedback locally does not rule out inference from transmitted parameters or auxiliary statistics~\citep{zhao2024leakage}. A compact update is not a privacy guarantee. Hence, correction updates need a defined protection unit, adversary, and privacy accounting. However, secure aggregation only limits the visibility, but does not validate the resulting behavior.

\textbf{Position 3: evaluate privacy and safety jointly when deciding what to share.} A rare correction may be unusually revealing and unusually important for avoiding harm. Protecting it may erase a safety signal; sharing it unprotected may expose the person or scene. Compare attack success at recovering goals or attributes, retention of safety-critical events, and receiver-side constraint violations at stated protection settings. Highly specific corrections may remain local, with only compatible, lower-risk evidence shared. 

\section{Risk-aware safety assurance}

For each correction-derived update, an inspectable record would state the source channel and context, the robot's immediate response, the proposed behavioral change and uncertainty, what leaves the client, and the receiver's acceptance tests and rollback condition. Such a record need not contain raw speech or video and cannot certify safety. It makes the reasons for sharing, withholding, or accepting an update explicit.

\textbf{Position 4: Risk and evidence-aware safety assurance.} Comparisons should keep the pretrained policy and safety scenarios fixed while varying modality, task context, embodiment, and sharing strategy. Local-only adaptation establishes what can be achieved without transfer, while indiscriminate and selective sharing expose both the benefits and the risks of exchanging corrections. Evaluation should consider not only task success, but also the frequency and severity of physical safety violations, human intervention effort, worst-case degradation at receiving clients, and information leakage under the assumed adversary. Because human intent may be ambiguous and simulation may fail to capture critical real-world interactions, correction transfer should not be treated as beneficial by default. Federated VLA adaptation should therefore require both a justified reason for sharing a correction and sufficient evidence that the recipient can incorporate it without unacceptable safety degradation. When such evidence is weak or uncertainty remains high, withholding the update should be considered a valid and desirable outcome.

\section{A falsifiable research agenda}

\textbf{Q1. Which corrections should be treated as equivalent?} In joint tasks, vary speech, gesture, teleoperation, and physical guidance across people, task states, and hazards. Test whether approaches distinguish vetoes, path changes, goals, preferences, and weak outcome reports, including uncertainty and conflicting signals. Measure whether each available channel produces an appropriate immediate defensive response, how well a collaborator anticipates a hold, yield, or restart, perceived safety separately from physical safety, and the human effort needed to intervene. Channel accessibility and communication failure should be reported, not averaged away.

\textbf{Q2. When does transfer regress safety?} Compare each recipient before and after a candidate update across embodiment, task, scene, and feedback distributions, including rare overrides. Measure physical violations, force, time to a safe state, interruption response, task success, and worst-recipient deterioration. Separate visual changes that should preserve an action from changes that require one, as studied for task variation in PAIR-VLA~\citep{peng2026pairvla}. Compare world-model predictions with observed cases, especially false acceptance, and test rejection, rollback, and unlearning of accepted updates. Risk-gated interactive learning offers relevant measures of supervisor burden~\citep{hoque2022thrifty}.

\textbf{Q3. What is revealed or lost by protecting a correction?} Test reconstruction and attribute inference on the actual shared information. Compare protection choices at explicit privacy settings, and check whether rare safety events still influence acceptable receivers. The evaluation should identify a safety event and threat model rather than collapse success, safety, human effort, and leakage into one score.

\textbf{Discriminating contrasts.} Hold the task and embodiment fixed while expressing the same intended constraint through different channels; then hold the channel fixed while changing the intended constraint. Transfer matched corrections across fixtures and embodiments, and vary the proportion of clients contributing rare overrides. Compare local-only, unrestricted federated, and scope-aware gated adaptation under each contrast. A useful gate should retain beneficial transfer while lowering false acceptance of unsafe updates; rejecting nearly everything would not support the value of federation. If ordinary aggregation preserves the same rare defensive behavior on every receiver, with comparable privacy protection and human effort, the case for the proposed restrictions weakens. These comparisons make the claim testable without presuming which method will win.

\section*{Acknowledgements}
This work was supported in part by the National Research Foundation, Prime Minister’s Office, Singapore under its Campus for Research Excellence and Technological Enterprise (CREATE) program. This work was partially supported by the Wallenberg AI, Autonomous Systems and Software Program (WASP) funded by the Knut and Alice Wallenberg Foundation via the WASP NEST project “Intelligent Cloud Robotics for Real-Time Manipulation at Scale.” 
\bibliography{references_vig}
\end{document}